\documentclass{article}
\PassOptionsToPackage{numbers,sort&compress}{natbib}
\usepackage[sglblindworkshop,final]{neurips_2026}
\usepackage[utf8]{inputenc}
\usepackage[T1]{fontenc}
\usepackage{amsmath,amssymb}
\usepackage{booktabs}
\usepackage{graphicx}
\usepackage{flafter}
\usepackage{microtype}
\usepackage{tikz}
\usetikzlibrary{arrows.meta,positioning}
\usepackage[hidelinks]{hyperref}
\usepackage{url}

\makeatletter
\renewcommand{\@noticestring}{Preprint.}
\makeatother

\title{SCB: SpeechConversationBench for Evaluating Multi-Turn Reasoning in Speech-to-Speech Model}
\author{%
Kanpat Vesessook\thanks{This work was primarily conducted during a 2024 internship at SCBX R\&D.}\\
SCBX R\&D\\
\texttt{t\_kanpat.v@scbx.com}
\And
Saksorn Ruangtanusak\\
SCB DataX, SCBX Group\\
\texttt{saksorn.ruangtanusak@data-x.ai}
}

\newcommand{\full}{\textsc{Full}}
\newcommand{\concat}{\textsc{Concat}}
\newcommand{\sharded}{\textsc{Sharded}}

\begin{document}
\maketitle

\begin{abstract}
Speech-to-speech systems must solve tasks whose requirements emerge across conversational turns. We introduce SpeechConversationBench (SCB), a focused evaluation of spoken mathematical reasoning using a pool of 103 sharded GSM8K problems. The framework compares the original problem delivered in one turn (\full), its concatenated information shards delivered together (\concat), and incremental spoken disclosure across turns (\sharded). We report final-answer accuracies for four commercial speech systems and LEGO, a proprietary speech pipeline developed internally by the SCBX Innovation Lab team, with explicit conversational context management. Relative to \concat, \sharded\ accuracy decreases by 5.0-25.3 percentage points across the four commercial systems. LEGO attains 77.5\% in all three conditions, compared with 76.6\% \sharded\ accuracy for GPT-4o Realtime. Both single-turn baselines are needed to distinguish sensitivity to reformulation from the challenges of incremental interaction.
\end{abstract}

\section{Introduction and Related Work}

Solving a complete spoken request does not establish that a system can solve the same task when relevant facts arrive over several turns. Mathematical word problems provide a focused test: success requires integrating incrementally disclosed quantities and constraints into a verifiable final answer.

\paragraph{Speech and dialogue evaluation.}
VoiceBench evaluates voice assistants across content, speaker, and environmental variations \citep{chen2024voicebenchbenchmarkingllmbasedvoice}, while Dynamic-SUPERB organizes diverse instruction-based speech tasks \citep{huang2023dynamic}. Multi-turn speech evaluation is also an established research direction. MTalk-Bench combines pairwise and rubric-based assessment across semantic, paralinguistic, and ambient-sound dimensions \citep{du2025mtalk}. Audio MultiChallenge evaluates natural spoken interactions involving memory, instruction retention, self-coherence, and spoken repairs \citep{gosai2025audio}. SpeechConversationBench complements these broader evaluations with a narrower comparison: numerical task completion under three presentations of corresponding problem information. It does not measure their full range of acoustic or interactional capabilities.

\paragraph{Incremental reasoning and memory.}
We adapt the \full/\concat/\sharded\ protocol of Laban et al.\ \citep{laban2025llmslostmultiturnconversation} to spoken mathematical tasks. LoCoMo instead evaluates memory over extended, multi-session conversations \citep{maharana2024locomo}; our setting concerns integrating the facts of one problem within an interaction.

\begin{samepage}
\paragraph{Architectural context.}
Moshi models full-duplex speech-text dialogue \citep{defossez2024moshi}, while MemGPT manages information across memory tiers \citep{packer2023memgpt}. Context-aware decoding strengthens adherence to spoken history \citep{lee2026awareness}. These approaches motivate our comparison of five speech systems, including the context-managing cascade LEGO; they do not validate LEGO itself.
\end{samepage}

\section{Evaluation Framework}

\subsection{Task and Input Conditions}

\begin{figure}[!ht]
\centering
\begin{tikzpicture}[
  font=\small,
  box/.style={draw=black!55,rounded corners=2pt,align=center,inner sep=5pt},
  arrow/.style={-{Stealth[length=4pt]},draw=black!65,line width=0.6pt}
]
\node[box,fill=black!4,text width=1.52cm,minimum height=1.0cm] (task) at (0,0) {GSM8K\\problem};
\node[box,fill=blue!5,text width=4.2cm] (full) at (3.65,1.03) {\full: original problem\\one spoken turn};
\node[box,fill=blue!5,text width=4.2cm] (concat) at (3.65,0) {\concat: combined shards\\one spoken turn};
\node[box,fill=blue!5,text width=4.2cm] (shard) at (3.65,-1.03) {\sharded: incremental shards\\multi-turn spoken exchange};
\node[box,fill=black!4,text width=1.52cm,minimum height=2.88cm] (model) at (7.2,0) {Speech\\system\\[4pt]evaluated in\\each condition};
\node[box,fill=green!6,text width=1.7cm,minimum height=1.1cm] (answer) at (9.62,0) {Final spoken\\answer\\[3pt]correctness};
\draw[arrow] (task.east) -- ++(0.25,0) |- (full.west);
\draw[arrow] (task.east) -- (concat.west);
\draw[arrow] (task.east) -- ++(0.25,0) |- (shard.west);
\draw[arrow] (full.east) -- (model.west |- full.east);
\draw[arrow] (concat.east) -- (model.west);
\draw[arrow] (shard.east) -- (model.west |- shard.east);
\draw[arrow] (model.east) -- (answer.west);
\end{tikzpicture}
\caption{Spoken evaluation conditions. \full\ and \concat\ present information together; \sharded\ distributes it across an exchange. Final spoken answers are assessed for numerical correctness.}
\label{fig:workflow}
\end{figure}
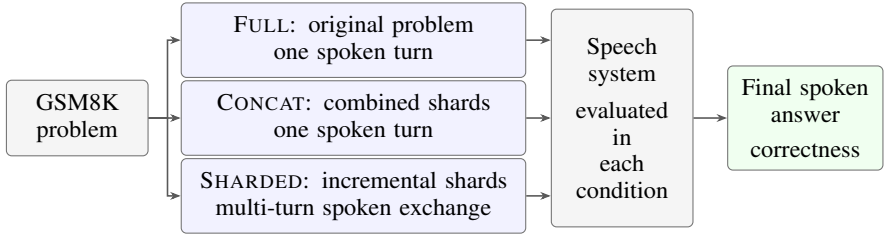

\begin{figure}[!ht]
\centering
\includegraphics[width=\linewidth]{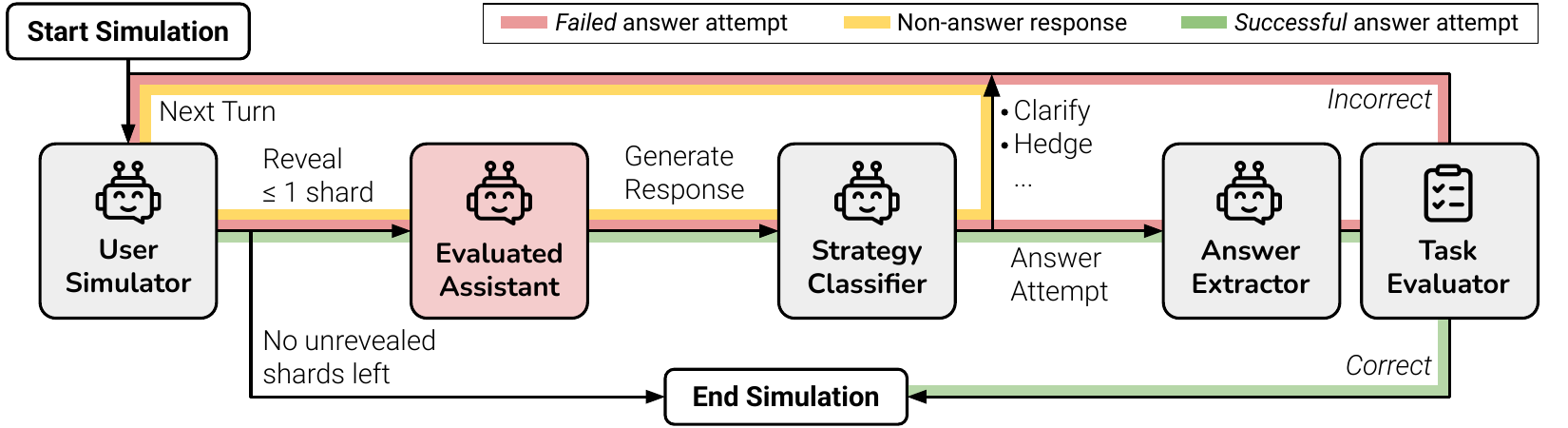}
\caption{Sharded-conversation simulator reproduced from Laban et al.\ \citep{laban2025llmslostmultiturnconversation}. This reference architecture motivates the spoken adaptation; it does not specify our audio-processing or scoring components.}
\label{fig:simulator}
\end{figure}

The problem pool contains 103 mathematical word problems from the sharded GSM8K resource of Laban et al.\ \citep{laban2025llmslostmultiturnconversation}. GSM8K contains grade-school problems requiring multi-step numerical reasoning \citep{cobbe2021trainingverifierssolvemath}. A shard is a partial statement of a problem's information; the shards jointly specify the task. Spoken prompts provide the input, and correctness of the final spoken answer is the evaluation target. Figure~\ref{fig:workflow} summarizes the conditions; Figure~\ref{fig:simulator} shows the original simulator.

In \full, the original complete problem is presented in one spoken turn. In \concat, the corresponding shards are combined and presented together in a single turn. In \sharded, information is disclosed incrementally through a multi-turn spoken exchange. Thus, \full\ and \concat\ differ in formulation, while \concat\ and \sharded\ differ in how the shard content is distributed through interaction. This follows the conceptual controls of the original text protocol \citep{laban2025llmslostmultiturnconversation} without assuming that its model settings transfer to speech.

The \concat\ baseline is especially useful because reformulation can itself change difficulty. Comparing \sharded\ only with \full\ can conflate that change with the effect of incremental disclosure. Nevertheless, \concat\ versus \sharded\ is not an isolation of memory alone: assistant responses, speech processing, and the evolving conversational trajectory may also contribute to the observed difference.

\subsection{Systems and Outcome Measures}

We evaluate GPT-4o Realtime, GPT-4o Mini Realtime, Gemini 2.5 Flash Live, and Gemini 2.5 Flash Preview Native Audio Dialog, alongside LEGO. These names identify the systems in the reported comparison; they do not establish exact API snapshots. LEGO is a proprietary pipeline developed internally by the SCBX Innovation Lab team. It combines automatic speech recognition (ASR), a large language model (LLM) with explicit context management, and text-to-speech (TTS) synthesis, with all component models self-hosted internally. The pipeline incorporates the Thai semantic end-of-turn detection method described by \citet{popit2025thai}. Its context mechanism summarizes conversational history and reintroduces relevant information at successive turns. This is a system-level comparison rather than a controlled test of any individual component.

Let $A_F$, $A_C$, and $A_S$ denote the reported percentages of correct final answers under \full, \concat, and \sharded. We summarize the multi-turn contrast by an absolute change and a normalized retention ratio:
\begin{equation}
\Delta_{S-C}=A_S-A_C,\qquad R_{S/C}=100\,\frac{A_S}{A_C}.
\label{eq:metrics}
\end{equation}
The change is measured in percentage points (pp); the ratio expresses \sharded\ accuracy as a percentage of \concat\ accuracy. A retention value of 100\% means equal aggregate accuracies, not that every problem has the same outcome. All derived quantities use the displayed accuracies and are rounded to one decimal place. The 103-problem pool is distinct from the per-condition trial denominators and repeat counts, which are not specified in the aggregate results. Consequently, we do not infer correct-answer counts or uncertainty intervals from that pool size.

\section{Results}

Table~\ref{tab:results} reports all five systems. Among the commercial systems, \sharded\ accuracy ranges from 46.6\% to 76.6\%, and every system scores lower in \sharded\ than in \concat. The absolute decreases are 5.0 pp for GPT-4o Realtime, 25.3 pp for GPT-4o Mini Realtime, 15.5 pp for Gemini 2.5 Flash Live, and 25.0 pp for Gemini 2.5 Flash Preview Native Audio Dialog. These correspond to retention ratios of 93.9\%, 67.4\%, 75.0\%, and 68.8\%, respectively. 

\begin{table}[!ht]
\centering
\caption{Final-answer accuracy (\%) under each input condition. $\Delta_{S-C}$ is the \sharded-minus-\concat\ change in percentage points; $R_{S/C}$ is aggregate accuracy retention (\%). Derived columns use the displayed accuracies. Bold marks the highest \sharded\ accuracy without implying significance.}
\label{tab:results}
\small
\begin{tabular}{@{}lrrrrr@{}}
\toprule
System & \full & \concat & \sharded & $\Delta_{S-C}$ & $R_{S/C}$\\
\midrule
GPT-4o Realtime & 70.0 & 81.6 & 76.6 & $-5.0$ & 93.9\\
GPT-4o Mini Realtime & 75.5 & 77.7 & 52.4 & $-25.3$ & 67.4\\
Gemini 2.5 Flash Live & 54.4 & 62.1 & 46.6 & $-15.5$ & 75.0\\
Gemini 2.5 Flash Preview$^{\dagger}$ & 85.0 & 80.0 & 55.0 & $-25.0$ & 68.8\\
\midrule
LEGO & 77.5 & 77.5 & \textbf{77.5} & $0.0$ & 100.0\\
\bottomrule
\end{tabular}

\smallskip
\parbox{\linewidth}{\footnotesize $^{\dagger}$Gemini 2.5 Flash Preview Native Audio Dialog.}
\end{table}

\paragraph{The baseline changes the interpretation.}
GPT-4o Realtime improves from 70.0\% in \full\ to 76.6\% in \sharded, even though it falls from 81.6\% in \concat. A \full-only comparison would therefore conceal its disadvantage relative to the concatenated-shard baseline. More generally, the \concat-minus-\full\ changes are $+11.6$, $+2.2$, $+7.7$, and $-5.0$ pp for the four commercial systems, respectively. Both single-turn controls are therefore informative; degradation is not universal across baseline choices.

\begin{samepage}
\paragraph{Single-turn rankings do not carry over.}
Gemini 2.5 Flash Preview Native Audio Dialog has the highest \full\ accuracy, 85.0\%, but achieves 55.0\% in \sharded. GPT-4o Realtime has lower \full\ accuracy yet the strongest \sharded\ result among the commercial systems. Selecting a system by single-turn accuracy can therefore favor a different system than selecting by multi-turn accuracy.

\end{samepage}

\paragraph{LEGO preserves aggregate accuracy.}
LEGO obtains 77.5\% in all three conditions and the highest observed \sharded\ score, exceeding GPT-4o Realtime by 0.9 pp. Its zero aggregate change makes explicit context management worth investigating, but does not show that summarization caused the result. The systems differ in their components, and no memory ablation is available. Equal aggregate scores could also conceal problems that become correct and others that become incorrect. We therefore interpret LEGO as a promising system-level observation, not proof that a cascade is generally superior to native audio modeling.

\clearpage
\section{Discussion, Limitations, and Conclusion}

\paragraph{What the comparison measures.}
The observed multi-turn gaps do not identify a failure mechanism. A wrong numerical answer can arise from misperception, loss or misuse of earlier information, arithmetic error, or answer rendering. Without intermediate transcripts and matched interventions, these explanations cannot be separated. The table establishes neither forgetting nor monotonic degradation with turn count.

\paragraph{Scope and reproducibility.}
This small mathematical evaluation does not measure open-domain or multilingual dialogue, interruptions, prosody, noise robustness, latency, speech naturalness, or memory-management cost. The aggregate record does not specify exact model snapshots, audio-generation settings, sampling parameters, per-condition denominators, repeat counts, or answer-extraction and scoring details. LEGO's component identities and memory-update implementation are also unspecified. These omissions limit reproduction and statistical interpretation.

\paragraph{Architectural interpretation.}
A matched text-only baseline is needed to quantify an audio-specific penalty. To test whether LEGO's context mechanism is responsible for its observed stability, the informative comparison would hold recognition, reasoning, and synthesis components fixed while varying context summarization. Problem-level outcomes and repeated runs would then support paired comparisons and uncertainty estimates. Such evidence would distinguish improvements in remembering information from improvements in using it, a distinction also raised by work on context-aware decoding \citep{lee2026awareness}. Broader spoken benchmarks remain necessary to assess whether gains in numerical correctness extend to natural interaction.

\paragraph{Conclusion.}
SpeechConversationBench compares spoken mathematical task completion across original, consolidated, and incrementally disclosed inputs. All four commercial systems lose accuracy from \concat\ to \sharded, whereas LEGO preserves its reported aggregate accuracy. The strongest supported lesson is methodological: retain both single-turn controls and report absolute task accuracy alongside the multi-turn change. 

\clearpage
\begin{ack}
We thank Monthol Charattrakool, Peerawat Rojratchadakorn, and Natthapath Rungseesiripak for their engineering contributions to LEGO. We also thank \href{https://th.linkedin.com/in/weerinchantaroje}{Weerin Chantaroje}, Head of Innovation Lab, SCBX, and Tutanon Sinthuprasith, Head of SCBX R\&D, for their leadership and support.
\end{ack}

\bibliographystyle{plainnat}
\bibliography{references}
\end{document}